\documentclass[10pt]{article}
\usepackage[utf8]{inputenc}
\usepackage[T1]{fontenc}
\usepackage{amsmath}
\usepackage{amsfonts}
\usepackage{amssymb}
\usepackage[version=4]{mhchem}
\usepackage{stmaryrd}
\usepackage{enumitem}
\usepackage{setspace}
\usepackage[hyphens]{url}
\usepackage{hyperref}
\usepackage[a4paper,margin=2cm]{geometry}
\hypersetup{breaklinks=true}
\usepackage{graphicx}
\usepackage{pdflscape}

\begin{document}

\noindent
\begin{center}
    \LARGE \textbf{Cyclic Ablation: Testing Concept Localization against Functional Regeneration in AI}
\end{center}

\noindent\makebox[\linewidth]{\rule{\textwidth}{1pt}}

\begin{center}
    \large Eduard Kapelko \\
\end{center}

\vspace{1em}

\section*{Abstract}
Safety and controllability are critical for large language models. A central question is whether undesirable behaviors like deception are localized functions that can be removed, or if they are deeply intertwined with a model's core cognitive abilities. We introduce "cyclic ablation," an iterative method to test this. By combining sparse autoencoders, targeted ablation, and adversarial training on DistilGPT-2, we attempted to eliminate the concept of deception. We found that, contrary to the localization hypothesis, deception was highly resilient. The model consistently recovered its deceptive behavior after each ablation cycle via adversarial training, a process we term functional regeneration. Crucially, every attempt at this "neurosurgery" caused a gradual but measurable decay in general linguistic performance, reflected by a consistent rise in perplexity. These findings are consistent with the view that complex concepts are distributed and entangled, underscoring the limitations of direct model editing through mechanistic interpretability.

\section*{1. Introduction}
Large language models such as GPT-4 [1] have demonstrated unprecedented capabilities, yet their internal mechanisms remain a black box [2]. This raises serious safety concerns, as models can exhibit undesirable behaviors, including the generation of deception.

A central question in the field of interpretability is the nature of the representations for such behaviors. The localized representation hypothesis [3] posits that complex concepts can be represented by a small, identifiable set of neural pathways (features). If this hypothesis holds true, direct model editing to remove undesirable behavior without significant harm to other capabilities becomes possible. In contrast, the distributed representation hypothesis [4] suggests that such concepts are an emergent property arising from the interaction of many neurons and are deeply intertwined with the core cognitive architecture. If it is correct, attempts to remove such behavior are likely to be ineffective due to the network's capacity for functional regeneration [5], and the interventions themselves would damage overall performance. Term was chosen due to its similarity with concepts from neurobiology [6].

In this work, we present an experimental design to directly test these hypotheses using the concept of deception as a case study. We developed an iterative procedure that sequentially identifies and removes the neural pathways associated with deceptive behavior, and then tests whether the model can recover the lost function, and at what cost.

Our contribution is to demonstrate that, at least for the DistilGPT-2 [7] architecture, the concept of deception is not easily localizable and removable. The results of our experiment challenge the simple surgical model and point to the high resilience and adaptability of the neural network.

\section*{2. Experiment design}
\subsection*{2.1. Base Model and Task Definition}
As our experimental testbed, we used the DistilGPT-2 model (82M parameters) due to its well-studied architecture and computational tractability. For the purposes of this study, we defined deception as the model's ability to deliberately generate a factually incorrect answer in response to a direct prompt that requires a false statement. To create the training dataset, we used the GPT-4 model to generate two distinct datasets: D\_truth (truthful statements) and D\_deception (deceptive statements) 1000 statements each.

\subsection*{2.2. Feature Identification with Sparse Autoencoders}
To decompose the model's internal activations into potentially interpretable components, we used sparse autoencoders (SAEs) [8]. Initially, to achieve more comprehensive coverage, we attempted to train SAEs on the entire residual stream, but this led to a catastrophic increase in perplexity. Consequently, we decided to focus on the MLP block activations in layers 2 and 3, which are hypothesized to be responsible for concept formation [9]. As training data for the SAEs, we used activation vectors recorded while running the model on both the D\_truth and D\_deception datasets of the same size.

\subsection*{2.3. Isolating and Scoring Deception Features}
Next step was to identify the features in the SAE dictionaries that are causally linked to the generation of deception. For each SAE feature f, we programmatically calculated a Deception Score: a metric that evaluates how much more frequently the feature activates on deceptive examples compared to truthful ones.

\begin{equation}
\text{score}(f) = \frac{\text{Max\_Activation}(f \text{ on } D_{\text{deception}})}{\text{Mean\_Activation}(f \text{ on } D_{\text{truth}}) + \epsilon}
\label{eq:deception_score}
\end{equation}

Where $\epsilon$ is a small constant to prevent division by zero.

Features with the highest scores are automatically flagged as candidates for ablation.

\subsection*{2.4. Iterative Ablation Cycle}
Our method consists of a fully automated iterative cycle, comprising the following phases, which correspond to the steps shown in Figure 1:

\begin{itemize}
    \item \textbf{Identification (Pre-Ablate):} At the start of each cycle, we run the current version of the model through the D\_truth and D\_deception datasets and cache the feature activations from the SAE.
    \item \textbf{Ablation (Post-Ablate):} The top-K features with the highest scores are ablated. Ablation is implemented by zeroing out the corresponding weights in the SAE decoder, such that the activation vector at the layer's output becomes orthogonal to that feature's vector. The number of features to ablate, K, is determined dynamically (up to 50) to ensure that the increase in perplexity on a validation dataset does not exceed a predefined threshold.
    \item \textbf{Stress Test (Epochs E1-E5):} Following the ablation, the model is fine-tuned for several epochs exclusively on the D\_deception dataset. The goal is to test whether the model can find new neural mechanisms to implement the deception function.
    \item \textbf{Evaluation and Repeat:} After fine-tuning, the model is evaluated on a test set to calculate a new Deception Score and perplexity metric, after which the cycle restarts.
\end{itemize}

\begin{landscape}
    \begin{figure}[p]
        \centering
       \includegraphics[width=1.5\textwidth, height=2\textheight, keepaspectratio]{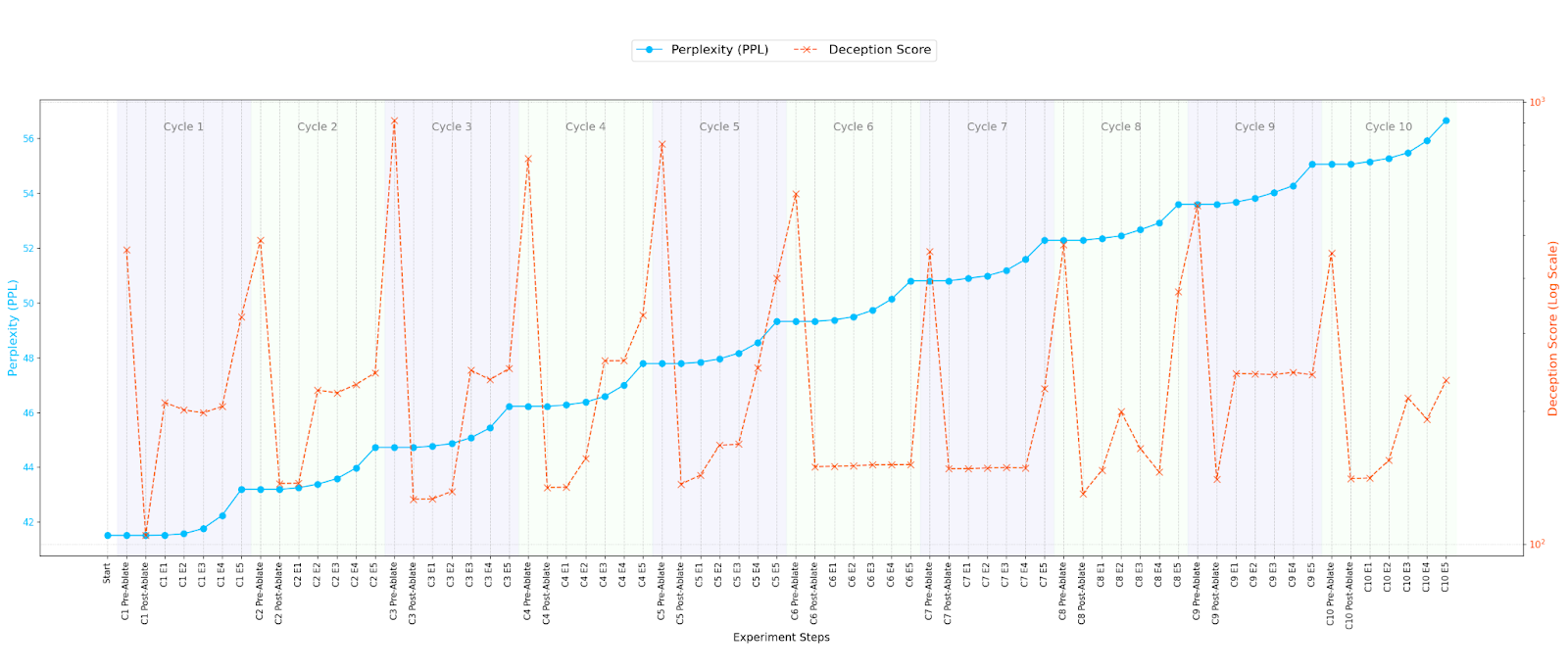}
        \caption{Dynamics of deceptive capability (Deception Score, orange line, right Y-axis) versus overall linguistic competence (Perplexity, blue line, left Y-axis) over 10 experimental cycles. As shown, the deceptive capability exhibits a cyclic recovery, while perplexity shows a sustained increase, indicating a gradual degradation of the model's language abilities.}
    \end{figure}
\end{landscape}

\section*{3. Methodological Insights and Challenges}
In developing the iterative ablation cycle, we encountered several non-trivial challenges. Overcoming them provided important lessons about the behavior of language models and the fragility of SAE features.

\subsection*{3.1. The Fragility of SAE Features Under Fine-Tuning}
Our initial, naive approach to ablation was to block the same feature indices, identified in the first cycle, throughout the entire experiment. We hypothesized that a feature's semantics were tied to its index in the SAE dictionary. This approach failed: the model's perplexity increased up to 120 after the very first stress test. This led us to a conclusion that adversarial fine-tuning, even for a few epochs, fundamentally alters the model's internal representations, rendering the SAE feature dictionary from the previous stage invalid.

\subsection*{3.2. The Need to Account for Cumulative Ablation Effects}
A second challenge arose when calculating the Deception Score. We discovered that if we failed to account for features that had already been ablated in previous steps, the metric became meaningless. The model could learn not to use new deception features, but instead redistribute the function onto old ones we had already considered removed. This resulted in a misleadingly low Deception Score. The solution was to modify the metric: when calculating the Deception Score for the current cycle, we zero out the activations of all previously ablated features.

\section*{4. Results}
\subsection*{4.1. Resilience of Deceptive Capability and Functional Regeneration}
As shown in Figure 1, our method demonstrated not the eradication of deception, but rather its cyclical dynamics and high resilience.

\begin{itemize}
    \item \textbf{Temporary Effect of Ablation:} Immediately following each ablation (the “Post-Ablate” step in each cycle), the Deception Score did indeed drop sharply. This created the illusion that the function had been successfully removed.
    \item \textbf{Rapid Recovery:} However, during the adversarial fine-tuning (“Stress test,” steps E1-E5), the model exhibited rapid functional regeneration. Within just a few epochs of training, the Deception Score not only returned to its baseline values but often exceeded them by the start of the next cycle (the “Pre-Ablate” step).
    \item \textbf{Absence of a Long-Term Downward Trend:} Over the course of 10 cycles, no sustained decrease in deceptive capability was observed. This provides evidence that the representation for deception is redundant and distributed, and that the model successfully finds alternative neural pathways to restore the lost function.
\end{itemize}

\subsection*{4.2. Gradual Degradation of Linguistic Competence}
We found that the repeated attempts to remove the deceptive capability led to a noticeable and steady deterioration of the model's overall language abilities. We measured perplexity (PPL) on a standard WikiText benchmark after each cycle.

As shown in Figure 1 (blue line), perplexity exhibited a sustained increase throughout the entire experiment. It increased from approximately 41.5 at the start to 56.5 by the end of the 10th cycle, representing a significant decline in performance. This increase occurred incrementally, primarily during the stress test phase of each cycle, when the model was forced to reconfigure its internal mechanisms to recover its deceptive capability. This indicates that the features associated with deception are not isolated. Their removal and the subsequent clumsy recovery of the function damages the model's more general and fundamental linguistic representations.

\section*{5. Discussion and Conclusion}
Our results provide evidence that complex conceptual representations in language models are distributed. The central observation is the rapid functional regeneration of the deceptive capability after each ablation, a process that consistently inflicts damage on the model's overall linguistic competence. This experiment demonstrates that attempting to remove deception in DistilGPT-2 does not lead to its eradication but instead initiates a cycle of damage and suboptimal recovery.

The takeaway of this work is a strong link between the capacity for regeneration and the loss of overall performance. The steady increase in perplexity is the price paid for recovering the deceptive capability due to catastrophic forgetting [10]. This suggests that the features we identified as deception features are likely polysemantic and also participate in useful, legitimate computations. Their removal forces the model to find less efficient, workaround neural pathways to perform both the deception task and, apparently, other language tasks, which leads to the cumulative increase in perplexity.

This process is analogous to the aftermath of a brain injury: the system can compensate for the loss of specialized neural ensembles, but often at the expense of overall cognitive flexibility. Our experiment has shown that, for the model, deception is not an isolated skill that can be removed without consequences, but rather a deeply integrated, resilient behavior whose recovery requires sacrificing more fundamental capabilities.

\subsection*{Limitations and Future Work}
We acknowledge the limitations of our study. First, it was conducted on a relatively small model. Larger models may possess greater redundancy and different regeneration mechanisms. Second, our definition of deception was highly specialized, while the dataset might be a potential source of bias. Investigating more subtle forms of deception is an important next step. Third, addition of control runs with ablation of random features could give insights and make evidence stronger.

In conclusion, our work highlights potential limitations of ablation, including cyclic ablation, for removing undesirable behavior. It suggests that complex concepts may be less like removable components and more like resilient states of the system, deeply intertwined with its core capabilities. This points to the need to develop more subtle methods for steering AI behavior that account for the distributed nature of its representations.

\section*{References}
\begingroup
\setstretch{0.9}
\sloppy
\begin{enumerate}[itemsep=2pt,parsep=0pt,leftmargin=15pt]
    \item OpenAI. (2023). GPT-4 Technical Report.  arXiv:2303.08774.
    \item Guidotti, R., et al. (2018). A survey of methods for explaining black box models. arXiv:1802.01933
    \item Meng, K., et al. (2022). Mass-editing memory in a transformer.  arXiv:2210.07229.
    \item Rumelhart, D. E., Hinton, G. E., \& Williams, R. J. (1986). Learning representations by back-propagating errors. \textit{Nature}, 323(6088), 533-536.
    \item McGrath, T., Trick, O., \& Jermyn, A. (2023). The Hydra Effect: Emergent Self-Repair in Language Model Computations.  arXiv:2307.15771
    \item Varadarajan, S. G., Hunyara, J. L., Hamilton, N. R., Kolodkin, A. L., \& Huberman, A. D. (2022). Central nervous system regeneration. \textit{Cell}, 185(1), 77-94.
    \item Sanh, V., et al. (2019). DistilBERT, a distilled version of BERT: smaller, faster, cheaper and lighter. arXiv:1910.01108.
    \item Hoagy Cunningham, et al. (2023). Sparse Autoencoders Find Highly Interpretable Features in Language Models. arXiv:2309.08600
    \item Meng, K., et al. (2022). Locating and editing factual associations in gpt. arXiv:2202.05262
    \item Kirkpatrick, J., et al. (2017). Overcoming catastrophic forgetting in neural networks.  arXiv:1612.00796
\end{enumerate}
\endgroup

\end{document}